\documentclass{article}

\usepackage{arxiv}

\usepackage[utf8]{inputenc} 
\usepackage[T1]{fontenc}    
\usepackage{hyperref}       
\usepackage{url}            
\usepackage{booktabs}       
\usepackage{amsfonts}       
\usepackage{nicefrac}       
\usepackage{microtype}      

\usepackage{natbib}
\usepackage{amsmath}
\usepackage{verbatim}
\usepackage{graphicx}
\usepackage{placeins}
\usepackage{xcolor}
\newcommand{\best}[1]{\textbf{#1}}

\renewcommand{\cite}[1]{\citep{#1}}

\newcommand{\colorann}[3]{}

\title{Zero-shot Slot Filling with DPR and RAG}
\author{
Michael Glass,
Gaetano Rossiello,
Alfio Gliozzo \\
IBM Research AI\\
mrglass@us.ibm.com, gaetano.rossiello@ibm.com, gliozzo@us.ibm.com
}

\begin{document}

\maketitle

\begin{abstract} 
The ability to automatically extract Knowledge Graphs (KG) from a given collection of documents is a long-standing problem in Artificial Intelligence. One way to assess this capability is through the task of slot filling.
Given an entity query in form of \textsc{[Entity, Slot, ?]}, a system is asked to `fill' the slot by generating or extracting the missing value from a relevant passage or passages. 
This capability is crucial to create systems for automatic knowledge base population, which is becoming in ever-increasing demand, especially in enterprise applications.
Recently, there has been a promising direction in evaluating language models in the same way we would evaluate knowledge bases, and the task of slot filling is the most suitable to this intent. The recent advancements in the field try to solve this task in an end-to-end fashion using retrieval-based language models. Models like Retrieval Augmented Generation (RAG) show surprisingly good performance without involving complex information extraction pipelines. 
However, the results achieved by these models on the two slot filling tasks in the KILT benchmark are still not at the level required by real-world information extraction systems.
In this paper, we describe several strategies we adopted to improve the retriever and the generator of RAG in order to make it a better slot filler. 
Our $KGI_0$ system\footnote{Our source code is available at: \url{https://github.com/IBM/retrieve-write-slot-filling}} reached the top-1 position on the KILT leaderboard on both T-REx and zsRE dataset with a large margin.

\end{abstract}

\section{Introduction}

A main barrier for adoption of KG technology for enterprise is the effort required to define the schema and populate enterprise specific relational data sources, such as KGs. In this work, we address this problem by exploring the use of zero-shot learning approaches for slot filling.

\begin{figure}[h!]
   \centering
   \includegraphics[width=0.9\linewidth]{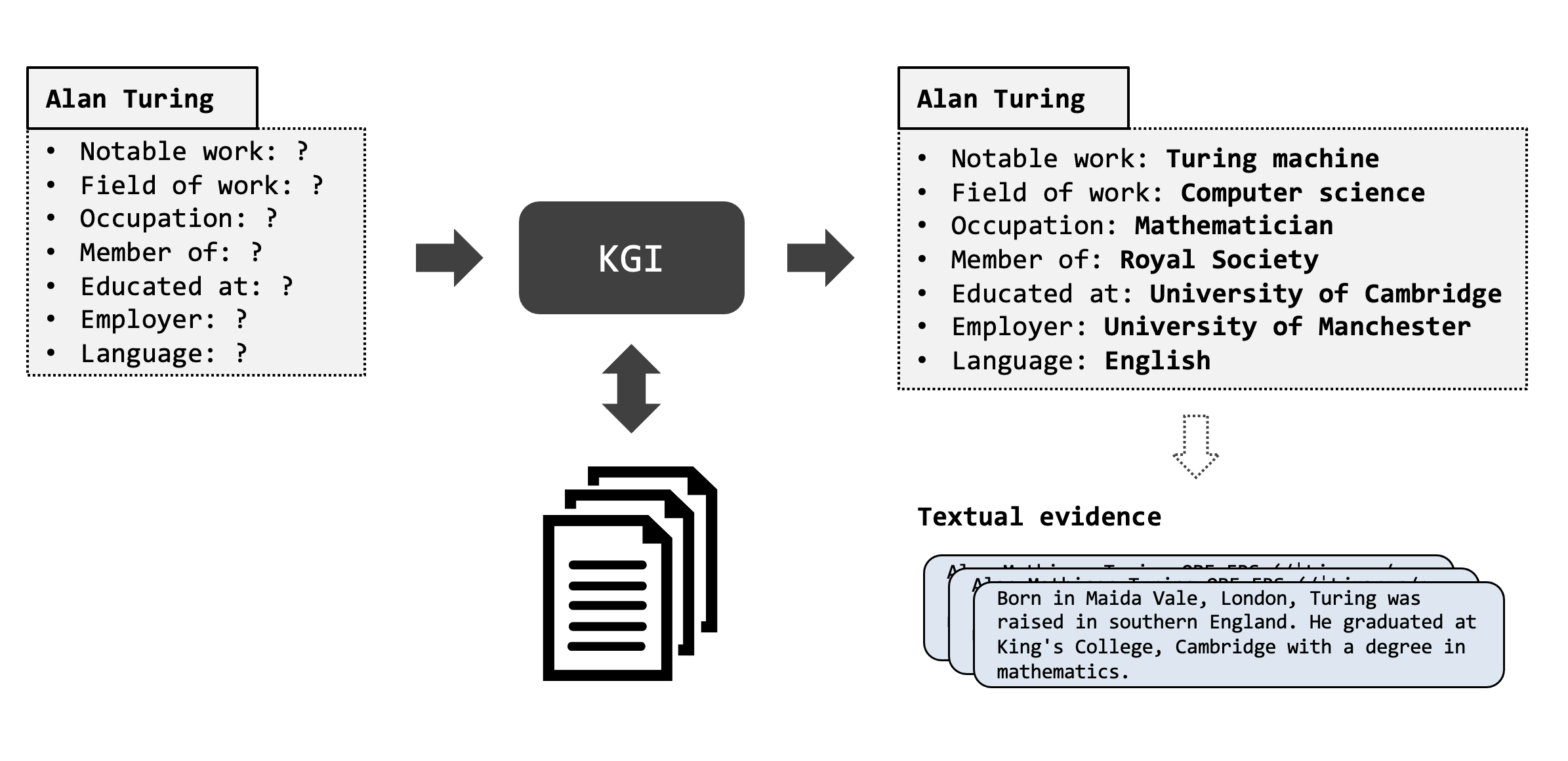}
   \caption{Knowledge Graph Induction from textual corpora}
   \label{fig.task}
\end{figure}

In the task of slot filling the goal is to identify a pre-determined set of relations for a given entity, and use them to populate infobox like structures. This can be done by exploring the occurrences of the input entity in the corpus and gathering information about its slot fillers from the context in which it it located.  Figure \ref{fig.task} illustrates the slot filling task. A slot filling system processes and indexes a corpus of documents, then when prompted with an entity and a number of relations, fills out an infobox and provides the evidence passages which explain the predictions. 

Over the past years, the proposed slot filling systems commonly involve complex pipelines for named entity recognition, entity co-reference resolution and relation extraction \cite{DBLP:conf/tac/EllisGFKSBS15}. In particular, the task of extracting relations between entities from text has been shown to be the weakest component of the chain. The community proposed different solutions to improve relation extraction performance, such as rule-based \cite{DBLP:conf/acl/AngeliPM15}, supervised \cite{DBLP:conf/emnlp/ZhangZCAM17}, or distantly supervised \cite{DBLP:conf/semweb/GlassGHMR18}. However, all these approaches require a considerable human effort in creating hand-craft rules, annotating training data, or building well-curated datasets for bootstrapping relation classifiers.

The use of language models as sources of knowledge \cite{DBLP:conf/emnlp/PetroniRRLBWM19, DBLP:conf/emnlp/RobertsRS20, DBLP:journals/corr/abs-2010-11967, DBLP:conf/akbc/PetroniLPRWM020}, has opened tasks such as zero-shot slot filling to pre-trained transformers.
The introduction of retrieval augmented language models such as RAG \cite{rag} and REALM \cite{realm} also permit providing textual provenance for the generated slot fillers.

A recently introduced suite of benchmarks, KILT (Knowledge Intensive Language Tasks) \cite{kilt}, standardizes two zero-shot slot filling tasks:
zsRE \cite{zsre} and T-REx \cite{trex}.  These tasks provide a competitive benchmark to drive advancements in slot filling.

One of the most interesting aspects of using pre-trained language models for zero-shot slot filling is the lower effort required for production deployment, which is a key feature for fast adaptation to new domains. However, the best performance achieved by the current retrieval-based models on the two slot filling tasks in KILT are still not satisfactory. This is mainly due to the lack of retrieval performance that affects the generation of the filler as well.

In this work, we propose a new slot filling specific training for both DPR and RAG. Furthermore, we observed that the RAG strategy of multiple sequence-to-sequence works better than the three passage concatenation in Mulit-DPR BART.
We implemented these ideas in our $KGI_0$ system, showing large gains on both T-REx (+24\% KILT-F1) and zsRE (+18\% KILT-F1) datasets.

\section{Related Work}\label{sec.related}



KILT was introduced with a number of baseline approaches. The best performing of these is RAG \cite{rag}. The model incorporates Dense Passage Retrieval (DPR) \cite{dpr} to first gather evidence passages for the query, then uses a model initialized from BART \cite{bart} to do sequence-to-sequence generation from each evidence passage concatenated with the query to generate the answer. In the baseline RAG approach only the query encoder and generation component are fine-tuned on the task. The passage encoder, trained on Natural Questions \cite{naturalquestions} is held fixed.
Interestingly, while it gives the best performance of the baselines tested on the task of producing slot fillers, its performance on the retrieval metrics is worse than BM25. This suggests that fine-tuning the entire retrieval component could be beneficial. 

In an effort to improve the retrieval performance, Multi-task DPR \cite{multidpr} used the multi-task training of the KILT suite of benchmarks to train the DPR passage and query encoder.  The top-3 passages returned by the resulting passage index were then combined into a single sequence with the query and a BART model was used to produce the answer.  This resulted in large gains in retrieval performance.

DensePhrases \cite{densephrases} is a different approach to knowledge intensive tasks with a short answer.
Rather than index passages which are then consumed by a reader or generator component, DensePhrases indexes the phrases in the corpus that can be potential answers to questions, or fillers for slots.  Each phrase is represented by the pair of its start and end token vectors from the final layer of a transformer initialized from SpanBERT-base-cased \cite{spanbert}. Question vectors come from the [CLS] token of two other transformers: one to be matched with the slot filler's start vector and one for the end vector. The start and end token vectors are indexed separately for maximum inner product search. Then at inference time the top-k start tokens are found for the question's start vector and the top-k end tokens are found for the question's end vector. These results are merged to find the top scoring phrase, which is then predicted as the slot filler.

GENRE \cite{genre} addresses the retrieval task in KILT slot filling by using a sequence-to-sequence transformer to generate the title of the Wikipedia page where the answer can be found.  This method can produce excellent scores for retrieval but does not address the problem of producing the slot filler. It is trained on BLINK \cite{blink} and all KILT tasks jointly.

\section{Knowledge Graph Induction}\label{sec.approach}

\begin{figure}[h!]
   \centering
   \includegraphics[width=0.6\linewidth]{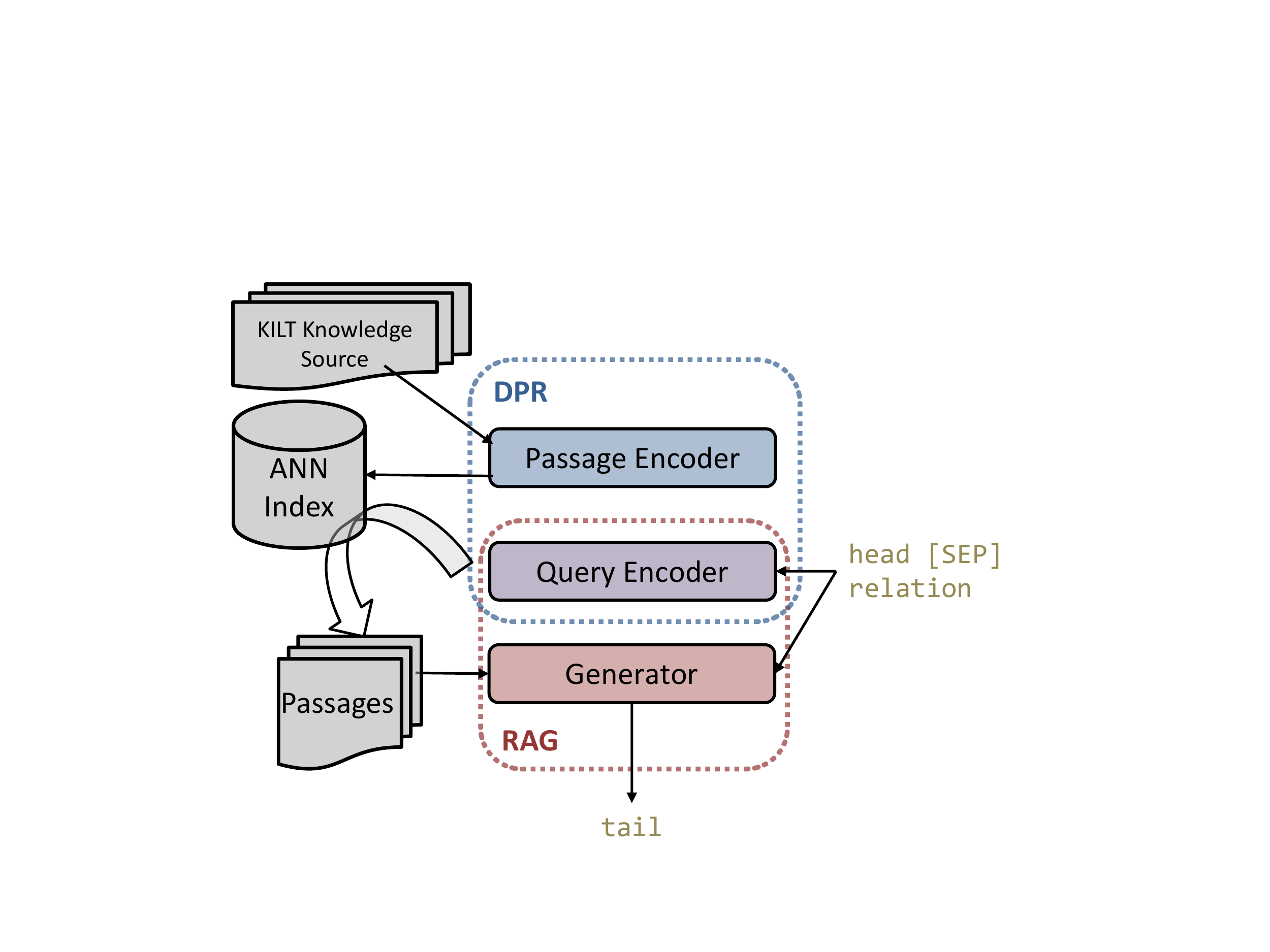}
   \caption{$KGI$ Architecture}
   \label{fig.arch}
\end{figure}

Figure \ref{fig.arch} shows Knowledge Graph Induction (KGI), our approach to zero-shot slot filling, combining a DPR model and RAG model, both trained for slot filling.  Due to the close connection between slot filling and open factoid question answering, we initialize our models from the Natural Questions\cite{naturalquestions}
trained models for DPR and RAG available from Hugging Face\footnote{\url{https://github.com/huggingface/transformers}}.
We then use a two phase training procedure: first we train the DPR model, i.e. both the query and context encoder, using the KILT provenance ground truth. Then we train the sequence-to-sequence generation and further train the query encoder using only the target tail entity as the objective.

Since the transformers for passage encoding and generation can accept a limited sequence length, we segment the documents of the KILT knowledge source (2019/08/01 Wikipedia snapshot) into passages. The ground truth provenance for the slot filling tasks is at the granularity of paragraphs, so we align our passage segmentation on paragraph boundaries when possible.  If two or more paragraphs are short enough to be combined, we combine them into a single passage and if a single paragraph is too long, we truncate it.

Our approach to DPR training for slot filling is a straightforward adaptation of the question answering training in the original DPR work \cite{dpr}. We first index the passages using a traditional keyword search engine, Anserini\footnote{\url{https://github.com/castorini/anserini}}. The head entity and the relation are used as a keyword query to find the top-k passages by BM25. Passages with overlapping paragraphs to the ground truth are excluded as well as passages that contain a correct answer. The remaining top ranked result is used as a hard negative for DPR training.

\begin{figure}[thb]
   \centering
   \includegraphics[width=0.8\linewidth]{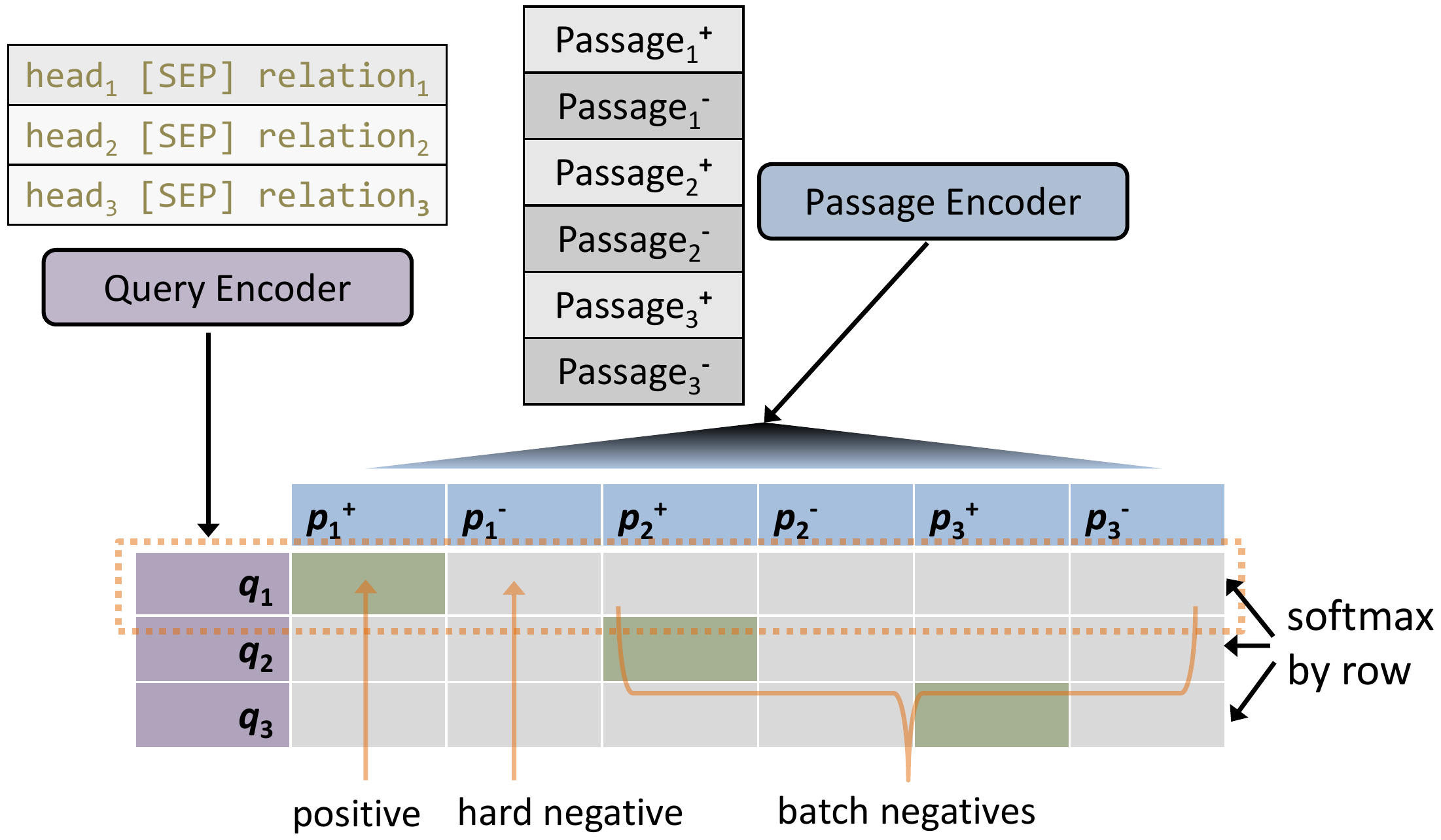}
   \caption{DPR Training}
   \label{fig.dpr}
\end{figure}


After locating a hard negative for each query, the DPR training data is a set of triples: query, positive passage (given by the KILT ground truth provenance) and our BM25 hard negative passage. Figure \ref{fig.dpr} shows the training process for DPR. For each batch of training triples, we encode the queries and passages independently. The passage and query encoders are BERT \cite{bert} models. Then we find the inner product of all queries with all passages. After applying a softmax to the score vector for each query, the loss is the negative log-likelihood for the positive passages.

Using the trained DPR passage encoder we generate vectors for the approximately 32 million passages in our segmentation of the KILT knowledge source. Though this is a computationally expensive step, it is easily parallelized.  
The passage-vectors are then indexed with an ANN (Approximate Nearest Neighbors) data structure, in this case HNSW (Hierarchical Navigable Small World)\cite{hnsw} using the open source FAISS \cite{faiss} library\footnote{\url{https://github.com/facebookresearch/faiss}}.
We use scalar quantization down to 8 bits to reduce the memory footprint.


The query encoder is also trained for slot filling alongside the passage encoder. We inject the trained query encoder into the RAG model for Natural Questions. Due to the loose coupling between the query encoder and the sequence-to-sequence generation of RAG, we can update the pre-trained model's query encoder without disrupting the quality of the generation.

\begin{figure}[thb]
   \centering
   \includegraphics[width=0.9\linewidth]{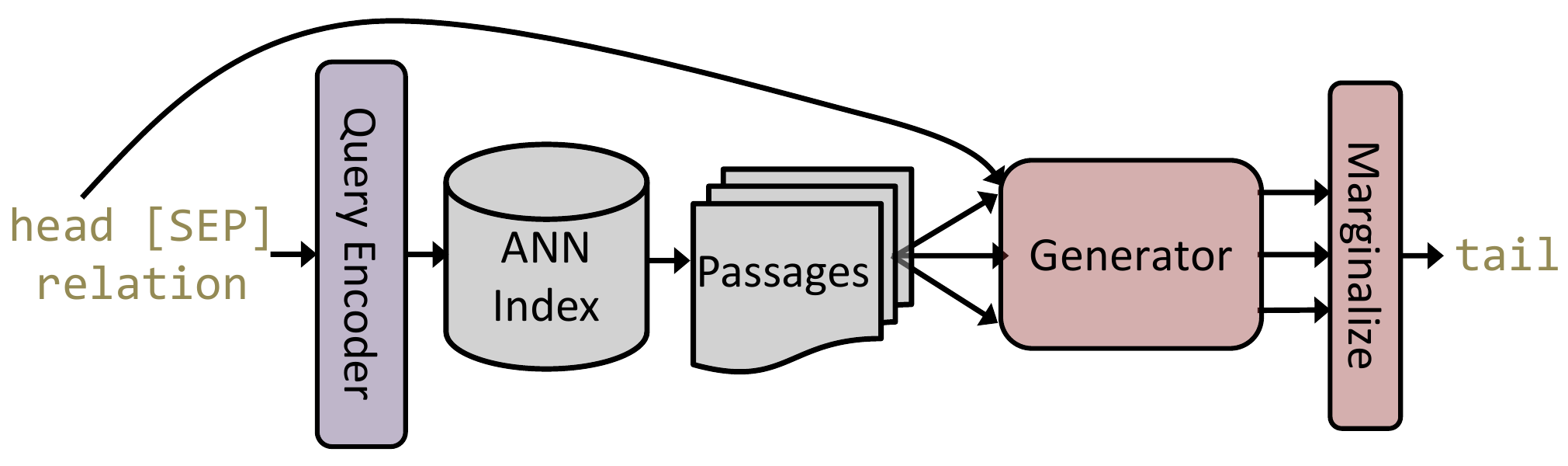}
   \caption{RAG Architecture}
   \label{fig.rag}
\end{figure}

Figure \ref{fig.rag} illustrates the architecture of RAG.
The RAG model is trained to predict the ground truth tail entity from the head and relation query.  First the query is encoded to a vector and relevant passages are retrieved from the ANN index.  The query is concatenated to each passage and the generator predicts a probability distribution over the possible next tokens for each sequence. These predictions are weighted according to the score between the query and passage - the inner product of the query vector and passage vector. The weighted probability distributions are then combined to give a single probability distribution for the next token.  Beam search is used to select the overall most likely tail entity.

Because the provenance used is at the level of passages but the evaluation is on page level retrieval, we retrieve up to twenty passages so that we typically get at least five documents for the Recall@5 metric.

We have not done hyperparameter tuning, instead using hyperparameters similar to the original works on training DPR and RAG. Table \ref{tbl.hypers} shows the hyperparameters used in our experiments. 

\begin{table}
\begin{center}
\begin{tabular}{rrr}
\textbf{Hyperparameter} & \textbf{DPR}  & \textbf{RAG} \\
\hline
learn rate  & 5e-5 & 3e-5 \\
batch size & 128 & 128 \\
epochs & 2 & 1* \\
warmup instances & 0 & 10000 \\
learning schedule & linear & triangular \\
max grad norm & 1 & 1 \\
weight decay & 0 & 0 \\
Adam epsilon & 1e-8 & 1e-8
\end{tabular}
\begin{small}
\begin{quote}
* Since the training set for T-REx is so large, we take only 500k instances.
\end{quote}
\end{small}
\end{center}
\caption{\label{tbl.hypers}$KGI_0$ hyperparameters}
\end{table}



\section{Experiments}

Table \ref{tbl.datasets} gives statistics on the two zero-shot slot filling datasets. While the T-REx dataset is larger by far in the number of instances, the training sets have a similar number of distinct relations. We use only 500 thousand training instances of T-REx in our experiments to increase the speed of experimentation.

\begin{table}
\begin{center}
\begin{tabular}{rrrrrrr}
 & \multicolumn{3}{c}{\textbf{Instances}} & \multicolumn{3}{c}{\textbf{Relations}}\\
\textbf{Dataset} & \textbf{Train}  & \textbf{Dev} & \textbf{Test} & \textbf{Train}  & \textbf{Dev} & \textbf{Test}\\
\hline
zsRE  &	147909 & 3724 & 4966 &	84 & 12 & 24 \\
T-REx &	2284168 & 5000 & 5000 &	106 & 104 & 104
\end{tabular}
\end{center}
\caption{\label{tbl.datasets}Zero-shot Slot Filling Dataset Sizes}
\end{table}

As an initial experiment we tried RAG with its default index of Wikipedia, distributed through Hugging Face. We refer to this as RAG-KKS, or RAG without the KILT Knowledge Source. Since the passages returned are not aligned to the KILT provenance ground truth, we do not report retrieval metrics for this experiment.  Motivated by the low retrieval performance reported for the RAG baseline by \citet{kilt}, we also experimented with replacing the DPR retrieval with simple BM25 (RAG+BM25). We provide the raw BM25 scores for the passages to the RAG model, to weight their impact in generation.  This provides a significant boost in performance.  Finally, we use the approach explained in Section \ref{sec.approach} to train both the DPR and RAG models. We call this system $KGI_0$, an initial knowledge graph induction system.

The metrics we report include accuracy and F1 on the slot filler, where F1 is based on the recall and precision of the tokens in the answer - allowing for partial credit on slot fillers.  Our systems (except for RAG-KKS) also provide provenance information for the top answer.  R-Precision and Recall@5 measure the quality of this provenance against the KILT ground truth provenance.  Finally, KILT-Accuracy and KILT-F1 are combined metrics that measure the accuracy and F1 of the slot filler \textit{only when the correct provenance is provided}. 
Table \ref{tbl.dev} gives our development set results.

\begin{table}
\begin{center}
\begin{tabular}{rrrrrrr}
\textbf{Method} & \textbf{R-Prec}  & \textbf{Recall@5} & \textbf{Accuracy} & \textbf{F1}  & \textbf{KILT-AC} & \textbf{KILT-F1}\\
\hline \multicolumn{7}{c}{\textbf{zsRE}} \\ \hline
RAG-KKS  &	 &  & 38.72\% & 46.94\% &  &  \\
RAG+BM25 &	58.86\% & 80.24\% & 45.73\% & 55.18\% & 36.14\% & 41.85\% \\
$KGI_0$ & 77.27\% & 96.37\% & 69.55\% & 97.66\% & 69.31\% & 76.83\% \\
\hline \multicolumn{7}{c}{\textbf{T-REx}} \\ \hline
RAG  &	 &  & 63.28\%  &	67.67\% &  &  \\
RAG+BM25 &	46.40\% & 67.31\% & 69.10\% & 73.11\% & 39.98\% & 41.21\% \\
$KGI_0$ & 61.30\%  & 71.18\% & 76.58\% & 80.27\% & 56.40\% & 57.70\% 
\end{tabular}
\end{center}
\caption{\label{tbl.dev}Dev. Set Performance for Various Retrieval Methods}
\end{table}

Table \ref{tbl.test} gives the test set performance of the top systems on the KILT leaderboard.  $KGI_0$ is our system, while DensePhrases, GENRE, Multi-DPR and RAG for KILT are explained briefly in Section \ref{sec.related}.  $KGI_0$ gains dramatically in slot filling accuracy over the previous best systems, with gains of over 10 percentage points in zsRE and even more in T-REx.  The combined metrics of KILT-AC and KILT-F1 show even larger gains, suggesting that the $KGI_0$ approach is effective at providing justifying evidence when generating the correct answer.  We achieve gains of 17 to 27 percentage points in KILT-AC.

\begin{table}[!th]
\begin{center}
\begin{tabular}{rrrrrrr}
\textbf{Method} & \textbf{R-Prec}  & \textbf{Recall@5} & \textbf{Accuracy} & \textbf{F1}  & \textbf{KILT-AC} & \textbf{KILT-F1}\\
\hline \multicolumn{7}{c}{\textbf{zsRE}} \\ \hline
$KGI_{0}$  & 94.18\% & 95.19\% & \best{68.97\%} & \best{74.47\%} & \best{68.32\%} & \best{73.45\%} \\
DensePhrases & 57.43\% & 60.47\% & 47.42\% & 54.75\% & 41.34\% & 46.79\% \\
GENRE & \best{95.81\%} & \best{97.83\%} & 0.02\% & 2.10\% & 0.00\% & 1.85\% \\
Multi-DPR & 80.91\% & 93.05\% & 57.95\% & 63.75\% & 50.64\% & 55.44\% \\
RAG (KILT organizers) & 53.73\% & 59.52\% & 44.74\% & 49.95\% & 36.83\% & 39.91\% \\
\hline \multicolumn{7}{c}{\textbf{T-REx}} \\ \hline
$KGI_{0}$  & 59.70\% & 70.38\% & \best{77.90\%} & \best{81.31\%} & \best{55.54\%} & \best{56.79\%} \\
DensePhrases & 37.62\% & 40.07\% & 53.90\% & 61.74\% & 27.84\% & 32.34\% \\
GENRE &	\best{79.42\%} & \best{85.33\%} & 0.10\% & 7.67\% & 0.04\% & 6.66\% \\
Multi-DPR & 69.46\% & 83.88\% & 0.00\% & 0.00\% & 0.00\% & 0.00\% \\
RAG (KILT organizers) &	28.68\% & 33.04\% & 59.20\% & 62.96\% & 23.12\% & 23.94\%
\end{tabular}
\end{center}
\caption{\label{tbl.test}KILT Leaderboard Top Systems}
\end{table}

\FloatBarrier

\section{Conclusion}

The $KGI_0$ system, combining slot filling specific training for both its DPR and RAG components, produces large gains in zero-shot slot-filling.
Our early experiments suggested the effectiveness of fine-tuning the retrieval component for the task, and highlighted the loose coupling of RAG's retrieval with its generation.
We find that DPR can be customized to the slot filling task and inserted into a pre-trained QA model for generation, to then be fine-tuned on the task.
Relative to Multi-DPR, we see the benefit of weighting passage importance by retrieval score and marginalizing over multiple generations, compared to the strategy of concatenating the top three passages and running a single sequence-to-sequence generation.
GENRE is still best in retrieval, suggesting that at least for a corpus such as Wikipedia, generating the title of the page can be very effective. 

\bibliographystyle{plainnat} 
\bibliography{main}

\end{document}